\documentclass[addpoints, 12pt]{article}
\usepackage{amsmath}
\usepackage{amssymb}
\usepackage{graphicx}
\usepackage[noadjust]{cite}
\usepackage{cite}

\begin{document}
\newcommand{\Rn}{\mathbb{R}^n}
\newcommand{\R}[1]{\mathbb{R}^{#1}}
\newcommand{\HRule}{\rule{\linewidth}{0.4mm}}

\begin{Large}
\begin{center}
\textbf{Notes on geometry of locomotion of 3-link body}
\end{center}
\end{Large}

\vspace{5pt}

\begin{center}
Sudin Kadam
\end{center}

\vspace{20pt}

We analyse the geometry of locomotion of 3-link mechanism inspired from the Purcell's swimmer at low Reynolds number, the simplest possible swimmer conceptualized in \cite{purcell1977life}. The literature has extensively analyzed the problem of plananr locomotion of the Purcell's swimmer \cite{avron2008geometric}, \cite{melli2006motion}, \cite{najafi2004simple}, \cite{or2009geometric}. \cite{hatton2013geometric} analyzes its locomotion problem in geometric framework, again for the planar case. The condition of being at low Reynold's number and slenderness of the links in mechanism leads us to a purely kinematic form of equations. Literature does not indicate any conditions of low Reynold's number theory for slender bodies which entail restriction to only planar locomotion. We extend the work to a more general, more challenging and more interesting generic 3 dimensional locomotion problem.

\begin{section}{Configuration Space}
The original form of Purcell's swimmer shown in fig. \ref{original_purcell01} has three links always in a common plane, the outer links were actuated through a hinge joint with the central link whose axes were perpendicular to the plane of mechanism

\begin{figure}[!htb]
\centering
\includegraphics[scale=.4]{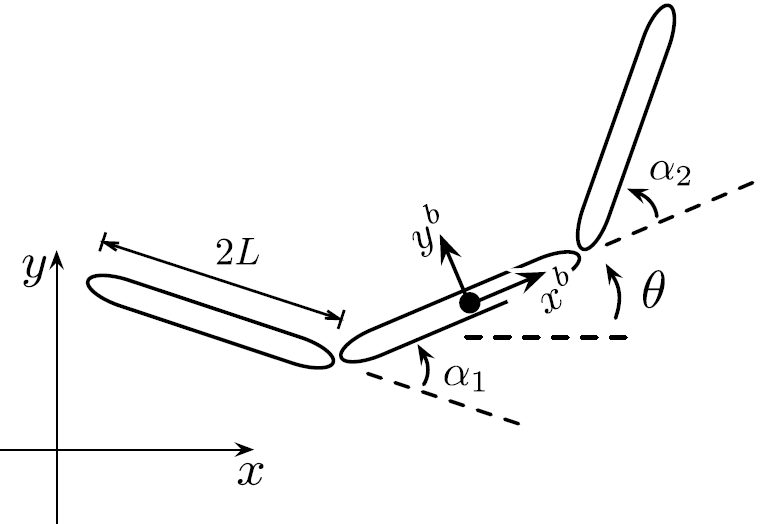}
\caption{Purcell's planar swimmer \cite{hatton2013geometric}}
\label{original_purcell01}
\end{figure}

We replace two hinge joints by ball joints allowing out-of-plane shape of the mechanism. This mechanism shall manoeuvre in Special Euclidean group ($SE3$). The translational position of the central (base) link denotes the position of its center of mass and is an element of $\mathbb{R}^3$, and its rotational position is an element of $SO(3)$. The outer 2 links being axisymmetric slender bodies, the rotation about the length direction of the body is immaterial, hence its orientation is an element of 2-sphere $\mathbb{S}^2$. Hence the configuration space is given by -

\begin{equation}
Q\:=\:\mathbb{R}^3\times SO(3) \times\mathbb{S}^2\times\mathbb{S}^2
\end{equation}
\end{section}

We represent the orientation of the outer links with respect to the base link through shape variables. The configuration space follows trivial principal fiber bundle topology \cite{kelly1995geometric}, with the position and orientation of the center link being the group variable $(g)$ which are the fibers over the shape space $\mathbb{S}^1\times\mathbb{S}^1$, formed by the orientation of the 2 limbs.

\begin{section}{Coordinate frames and transformations}

Fig \ref{fig:1 3Link arbit} shows an arbitrary position of the system along with 3 coordinate frames corresponding to each link, with their origin at the center of mass of the respective link. The reference configuration is the one in which shape variables are zero, i.e. all the 3 coordinate frames are aligned to each other. The orientation of the outer links is parametrized using usual latitude-longitude parametrization of $\mathbb{S}^2$. The orientation of the outer link in any arbitrary position can be achieved by successive rotations, first along base link's z-axis by $\psi$, followed by the rotation about the y-axis of the link being rotated by $\theta$. The usual rotation matrices for rotation about each of z and y axes are as follows- 

\begin{figure}[!htb]
\centering
\includegraphics[scale=.7]{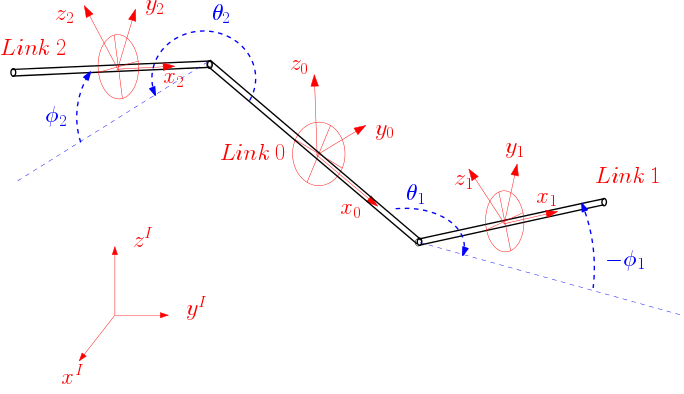}
\caption{Basic Configuration}
\label{fig:1 3Link arbit}
\end{figure}

\begin{equation}\nonumber
R_y(\theta)=\begin{pmatrix}
    \cos \theta   & 0 	& -\sin \theta \\
    0      		& 1 	& 0\\
    \sin \theta	& 0 	& \cos \theta
\end{pmatrix}, \qquad    
    R_z (\phi)=\begin{pmatrix}
    \cos \phi       & -\sin \phi 	& 0 \\
    \sin \phi       & \cos \phi 	& 0\\
    0				& 0 	& 1
\end{pmatrix}
\end{equation}

Orientation of any of the outer links can be represented with respect to the base link orientation using composite rotation as follows 

\begin{equation}
R_y(\phi)R_z(\theta)=\begin{pmatrix}
    \cos \theta \cos \phi   & -\sin \theta \cos \phi 	& -\sin \phi \\
    \sin \theta      		& \cos \theta 	& 0\\
   -\sin \phi \cos \phi	& \sin \phi \sin \theta 	& \cos \phi
\end{pmatrix}
\end{equation}

$\theta_1, \phi_1, \theta_2, \phi_2$ define the coordinates of the shape space. We can write the kinematics of the i'th link in terms of the velocity of the base link and the joint velocities $\dot{\theta}_1, \dot{\phi}_1, \dot{\theta}_2, \dot{\phi}_2$. Velocity of i'th link in its own body frame is an element of Lie algebra, which is isomorphic to $\mathbb{R}^n$. In our case the this is $se(3)$, which is represented as a generalized velocity vector, an element of $\mathbb{R}^6$ as $\xi_i\:=\: \begin{bmatrix}
    v_x, \: v_y,\: v_z,\: \omega_x,\: \omega_y,\: \omega_z \end{bmatrix}^T$

\end{section}

\begin{section}{Fluid forces} 

At very low Reynolds numbers, viscous drag forces dominate
the fluid dynamics of swimming and any inertial effects
are immediately damped out. This effect has two consequences \cite{hatton2013geometric}. First, the drag forces on the swimmer are linear functions of the body and shape velocities. Second, the net drag forces and moments on an isolated system interacting with the surrounding fluid go to zero: if the swimmer were to move with any velocity other than that dictated by force equilibrium, the large viscous forces would almost instantaneously remove this excess velocity, returning the system to the equilibrium velocity.

We model three-link swimmer with links as slender members leading to fluid forces according to Cox theory \cite{cox1970motion}. We regard the flows around each link as independent, according to resistive force theory \cite{tam2007optimal}. The drag forces and moments on the ith link are based on a principle of lateral drag coefficients being larger than those in the longitudinal direction, with a maximum ratio of 2 : 1 in the limit of an infinitesimally thin member. The moment around the center of the link is found by taking the lateral drag forces as linearly distributed along the link according to its rotational velocity. Thus, with $k$ as the differential viscous drag
constant and $2L$ as the length of each link, forces and moments on the i'th link are as follows,

\begin{equation}
F_{i,x}\:=\:\int_{-L}^L \frac{1}{2}k\xi_{i,x} dL \:=\: k \xi_{i,x} L
\end{equation}

\begin{equation}
F_{i,y}\:=\:\int_{-L}^L k\xi_{i,y} dl \:=\: 2 k \xi_{i,y} L
\end{equation}

\begin{equation}
F_{i,z}\:=\:\int_{-L}^L k\xi_{i,z} dl \:=\: 2 k \xi_{i,z} L
\end{equation}

\begin{equation}
M_{i}\:=\:\int_{-L}^L [0,\: \xi_{i,\theta_y},\: \xi_{i,\theta_z}]^T kl^2 dl \:=\: \frac{2}{3} [0,\: \xi_{i,\theta_y},\: \xi_{i,\theta_z}]^T k L^3
\end{equation}

Thus the total force on each link can be written as the linear combination of its body velocities -

\begin{equation}\label{force_linear_combination}
F_i\:=\:\begin{bmatrix}
kL & 0 & 0 & 0 & 0 & 0 \\
0 & 2kL & 0 & 0 & 0 & 0 \\
0 & 0 & 2kL & 0 & 0 & 0 \\
0 & 0 & 0 & 0 & 0 & 0 \\
0 & 0 & 0 & 0 & \frac{2}{3} kL^3 & 0 \\
0 & 0 & 0 & 0 & 0 & \frac{2}{3} kL^3
\end{bmatrix}\begin{pmatrix}
    \xi_{i,x} \\
    \xi_{i,y} \\
    \xi_{i,z}	\\
    \xi_{i,\theta_x} \\
    \xi_{i,\theta_y} \\
    \xi_{i,\theta_z} \\
\end{pmatrix}
\end{equation}

Thus,
\begin{equation}\label{Force_ith_link}
F_i\:=\:A\xi_i
\end{equation}

\end{section}

\section{Construction of connection form}
A mechanical system is kinematic \cite{shammas2007geometric} when the first derivative of its state vector is linearly dependent on the control inputs, i.e. $\dot{q}=A(q)u$, where $q$ is a state vector, $A(q)$ is a matrix and $u$ is the input vector. Purely kinematic systems is a type of kinematic system, that have as many independent non-holonomic constraints as the dimension of the system's fiber space. The motion of such systems with a configuration $q = (g,r) \in Q$ can be described by the reconstruction equation, $\xi = A(g,r)\dot{r}$, where $\xi$ is the body representation of fiber velocity, $A$ is a matrix that depends on the configuration $q$ and we treat $r$ as input vector.

In order to derive the force on each link, from eqn.  \ref{Force_ith_link}, we see that we need to find the body velocity of each link in terms of body velocity of the base link and the shape velocities, which is done through appropriate frame transformations as below for link 1 -

\begin{align*}
\xi_1\:&=\: \begin{bmatrix}
    R_y (\theta)R_z (\theta) & R_y (\theta)R_z (\theta)
\end{bmatrix}
\begin{pmatrix}
    v_x \\
    v_y \\
    v_z	\\
    \omega_x \\ 
    \omega_y \\ 
    \omega_z
\end{pmatrix} \\
&\qquad \qquad + \begin{pmatrix}
    \begin{bmatrix}
    	\dot{\theta_1} \sin \phi_1 \\ 
	    \dot{\phi_1} \\ 
	    \dot{\theta_1} \cos \phi_1
    \end{bmatrix} \times 
    \begin{bmatrix}
    	L\\
    	0\\
    	0
    \end{bmatrix} + 
    \begin{bmatrix}
    	\omega_x\\
    	\omega_y\\
    	\omega_z
    \end{bmatrix} \times 
    \begin{bmatrix}
    	L+L \cos \phi \cos \theta\\
    	L \cos \phi \sin \theta\\
    	L \sin \phi
    \end{bmatrix}\\
    \begin{bmatrix}
    \dot{\theta_1} \sin \phi_1 \\ 
    \dot{\phi_1} \\ 
    \dot{\theta_1} \cos \phi_1    
    \end{bmatrix}
\end{pmatrix}  \\
&= \left[
\begin{array}{c|c|c}
R_y (\theta)R_z (\theta) &	\begin{bmatrix}
    	L+L \cos \phi \cos \theta\\
    	L \cos \phi \sin \theta\\
    	L \sin \phi
    \end{bmatrix}^\times & \begin{bmatrix}
    	0 & L \\
    	-L \cos \phi_2 0\\
    	0 & 0
    \end{bmatrix}\\
\hline
\large 0_{3\times3} &	R_y (\theta)R_z (\theta) & \begin{bmatrix}
    	\sin \phi_1 & 0 \\
    	-L\cos \phi_1 0\\
    	\cos \phi_1 & 0
    \end{bmatrix}
\end{array}
\right]\begin{pmatrix}
    v_x \\
    v_y \\
    v_z	\\
    \omega_x \\ 
    \omega_y \\ 
    \omega_z \\
    \dot{\theta} \\
    \dot{\phi}
\end{pmatrix}
\end{align*}

Similar expression can be derived for the second link leading us to the following form of the velocity of each of links 0,1 and 2 -

\begin{equation}
\xi_0=\begin{pmatrix}
    \xi_{i,x} \\
    \xi_{i,y} \\
    \xi_{i,z}	\\
    \xi_{i,\theta_x} \\
    \xi_{i,\theta_y} \\
    \xi_{i,\theta_z}
\end{pmatrix}, \qquad \xi_1=B_1\begin{pmatrix}
    \xi_0 \\
    \dot{\theta_1} \\
    \dot{\phi_1}\end{pmatrix}, \qquad \xi_2=B_2\begin{pmatrix}
    \xi_0 \\
    \dot{\theta_2} \\
    \dot{\phi_2}    
\end{pmatrix}
\end{equation}

Combination of 2 effects mentioned before due to low Reynold's number results in the equations of motion taking on the form of a kinematic reconstruction equation. In the following section we show that the 3-link swimmer is a purely kinematic system and derive the reconstruction equation in coordinates. From fluid force equations \ref{force_linear_combination} we see that the force on each link is a linear combination of the velocity of the link in its own frame. Hence the force on each of the 3 links is - 

\begin{equation}
F_0\:=\:A\xi_0, \qquad F_1\:=\:AB_1 \begin{pmatrix}
    \xi_0 \\
    \dot{\theta_1} \\
    \dot{\phi_1}\end{pmatrix}, \qquad F_2\:=\:AB_2 \begin{pmatrix}
    \xi_0 \\
    \dot{\theta_1} \\
    \dot{\phi_1}
    \end{pmatrix}
\end{equation}

The summation of all the forces gives us the resultant force acting on the system. But we note that eqn (\ref{Force_ith_link}) gives us the force with respect to the frame of the same link. Hence before summing us we transform the forces to the frame associated with base link.

\begin{equation}
F\:=\:F_0 + T_1F_1 + T_2F_2
\end{equation}

Where the transformation matrices corresponding to outer links are -

\begin{equation}
T_1=\begin{bmatrix}
R_z (\theta)R_y (\phi) & \vline &	0\\
\hline
\large \begin{bmatrix}
    	L+L \cos \phi \cos \theta\\
    	L \cos \phi \sin \theta\\
    	L \sin \phi
    \end{bmatrix}^\times R_z (\theta)R_y (\theta) & \vline &	R_y (\theta)R_z (\theta)
\end{bmatrix}
\end{equation}

\begin{equation}
T_2=\begin{bmatrix}
R_z (\theta)R_y (\phi) & \vline &	0\\
\hline
\large \begin{bmatrix}
    	L+L \cos \phi \cos \theta\\
    	L \cos \phi \sin \theta\\
    	L \sin \phi
    \end{bmatrix}^\times R_y (\phi)R_z (\theta) & \vline &	R_y (\phi)R_z (\theta)
\end{bmatrix}
\end{equation}

The purpose is to write the equation in the form of a reconstruction equation. We can split these equation by writing matrices in terms of their block format.

\begin{align*}
F&=A\xi_0 + T_1 AB_1 \begin{pmatrix}
    \xi_0 \\
    \dot{\theta_1} \\
    \dot{\phi_1}\end{pmatrix} + T_2AB_2 \begin{pmatrix}
    \xi_0 \\
    \dot{\theta_2} \\
    \dot{\phi_2}\end{pmatrix} \\
   &= A\xi_0 + [[T_1 AB_1]_{6\times6} \:\: [T_1 AB_1]_{6\times2}] \begin{pmatrix}
    \xi_0 \\
    \dot{\theta_1} \\
    \dot{\phi_1}\end{pmatrix} + [[T_2AB_2]_{6\times6} \:\: [T_2AB_2]_{6\times2}] \begin{pmatrix}
    \xi_0 \\
    \dot{\theta_2} \\
    \dot{\phi_2}\end{pmatrix}
\end{align*}

Consequence of being at low Reynolds number is that the net forces and moments on an isolated system should be zero, which leads to following equation -

\begin{equation}
0\:=\:A\xi_0 + [T_1 A B_1]_{6\times6}\xi_0 +[T_1 A B_1]_{6\times2} \begin{pmatrix}
    \dot{\theta_1} \\
    \dot{\phi_1}
    \end{pmatrix} + + [T_2 A B_2]_{6\times6}\xi_0 + [T_2 A B_2]_{6\times2} \begin{pmatrix}
    \dot{\theta_2} \\
    \dot{\phi_2}
    \end{pmatrix}
\end{equation}

Finally we get a reconstruction equation in desired form as follows -

\begin{equation}
\xi_0 = \omega_1^{-1}\omega_2 \dot{r}
\end{equation}

where $\dot{r} = \begin{bmatrix}
	\dot{\theta_1},\: \dot{\phi_1},\: \dot{\theta_2},\:   \dot{\phi_2}    \end{bmatrix}^T$ is the vector of shape velocities $\omega_1$ and $\omega_2$ matrices are given as follows -

\begin{align*}
\omega_1 &=  [A+[T_1AB_1]+[T_2AB_2]]_{6\times6} \:\: \text{and} \\ \omega_2 &=  [[T_1 A B_1]_{6\times2}\:\:\:\: [T_2 A B_2]_{6\times2}]_{6\times4}
\end{align*}

\section{Pointers for next work related to connection vector field}
\begin{itemize}
\item \cite{hatton2008connection}, \cite{alouges2013self}, \cite{hatton2009approximating} use the concept of connection vector field, which is the vector field formed by the rows of the connection form. The inspection of this vector field gives insights into the effect of shape velocities on components of group velocities. The curvature of the connection vector field facilitates in gait analysis.

\item The \cite{hatton2011geometric} does the controllability and geometric maneuverability for the planar case.

\item \cite{kanso2005optimal} analyses optimality of shape change for generating desired global motion using nonlinear optimization problem.

\end{itemize}

\bibliographystyle{ieeetran}
\bibliography{bib1}

\end{document}